# SignBank+: Preparing a Multilingual Sign Language Dataset for Machine Translation Using Large Language Models

**Amit Moryossef**[†‡], **Zifan Jiang**[‡]

Bar-Ilan University[†], University of Zürich[‡]

amitmoryossef@gmail.com, jiang@cl.uzh.ch

## Abstract

We introduce SignBank+, a clean version of the SignBank dataset, optimized for machine translation between spoken language text and SignWriting, a phonetic sign language writing system. In addition to previous work that employs complex factorization techniques to enable translation between text and SignWriting, we show that a traditional text-to-text translation approach performs equally effectively on the cleaned SignBank+ dataset. Our evaluation results indicate that models trained on SignBank+ surpass those on the original dataset, establishing a new benchmark for SignWriting-based sign language translation and providing an open resource for future research.

## 1. Introduction

Sign Language serves as an indispensable mode of communication for the deaf. Unfortunately, the available methods for translating between signed and spoken languages, have been limited in scope and effectiveness (Müller et al., 2022, 2023). The main objective of this research is to explore technological advancements that can enhance sign language machine translation, focusing on the cleaning and enrichment of an existing sign language dataset, *SignBank*[1], a collection of *puddles*[2], covering a range of domains from bible books, and literature to lexicons and encyclopedias.

Jiang et al. (2023) present an approach to translating from/to SignWriting (Sutton, 1990) (a phonetic sign language writing system) through specialized parsing and factored machine translation (Koehn and Hoang, 2007; Garcia-Martinez et al., 2016) techniques. This work aims to build upon their foundation by:

1. Undertaking an efficient data cleaning process and extending the dataset they utilized.

2. Reverting to a simple text-to-text translation mechanism omitting any factorization with the cleaned dataset.

The motivation driving this study is twofold: First, a meticulously curated dataset will enhance the accuracy and reliability of translation models. Second, a simplified machine translation mechanism makes model training and deployment easier.

To validate our claims, we compare the translation quality of signed-to-spoken translation using the original and cleaned data. We show that with our newly cleaned data, we can train standard machine translation models with significantly improved quality over the original data. We share our data and code openly under CC-BY-4.0 (available at https://github.com/sign-language-processing/signbank-plus).

## 2. Background

This work only concerns machine translation between signed and spoken languages where both the input and the output are represented as discrete tokens (or, text).

### 2.1. Signed-to-Spoken

Jiang et al. (2023) explore text-to-text sign-to-spoken language translation, with SignWriting as the sign language notation system. Despite SignWriting usually being represented in 2D, they use the 1D Formal SignWriting specification and propose a neural factored machine translation approach to encode sequences of SignWriting graphemes as well as their positions in the 2D space. They verify the proposed approach on the SignBank dataset in both a bilingual setup (American Sign Language to English) and two multilingual setups (4 and 21 language pairs, respectively). They apply several low-resource machine translation techniques used to improve spoken language translation to improve the performance of sign language translation similarly. Their findings validate the use of an intermediate text representation for signed language translation and pave the way for including sign language translation in natural language processing research.

---

[1] https://www.signbank.org/signpuddle/

[2] Each puddle typically represents sign language data for a specific language or dialect.

## 2.2. Spoken-to-Signed

Jiang et al. (2023) also explore the reverse translation direction, i.e., text to SignWriting translation. They conduct experiments under a same condition of their multilingual SignWriting to text (4 language pairs) experiment, and again propose a neural factored machine translation approach to decode the graphemes and their position separately. They borrow BLEU from spoken language translation to evaluate the predicted graphemes and mean absolute error to evaluate the positional numbers.

Walsh et al. (2022) explore Text to HamNoSys (Prillwitz and Zienert, 1990) (T2H) translation, with HamNoSys as the target sign language notation system. They experiment with direct T2H and Text to Gloss to HamNoSys (T2G2H) on a subset of the data from the MEINE DGS dataset (Hanke et al., 2020), where all glosses are mapped to HamNoSys by a dictionary lookup. They find that direct T2H translation results in higher BLEU (it still needs to be clarified how well BLEU represents the quality of HamNoSys translations, though). They encode HamNoSys with BPE (Sennrich et al., 2016), outperforming character-level and word-level tokenization. They also leverage BERT (Devlin et al., 2019) to create better sentence-level embeddings and use HamNoSys to extract the hand shapes of a sign as additional supervision.

## 2.3. Machine Translation Frameworks

Machine translation has witnessed substantial advancements in recent years, both in terms of model architectures and frameworks that facilitate their training and deployment. When it comes to text-to-text translation, several open-source platforms have emerged, leading to the democratization of machine translation technology.

Prominent machine translation frameworks include *OpenNMT* (Klein et al., 2017), *Sockeye* (Hieber et al., 2017, 2020), Joey NMT (Kreutzer et al., 2019), and *Fairseq* (Ott et al., 2019). They are all widely renowned for simplicity, efficiency, and emphasis on performance, promoting rapid prototyping and thus becoming a popular choice among machine translation researchers.

## 3. Data

The SignBank dataset, while invaluable, includes numerous inconsistencies and imperfections arising from its community-contributed nature, which poses challenges for machine translation. Multiple non-parallel textual entries were associated with singular signing sequences. For instance, while some entries indicate chapter and page numbers from a book, the actual text is missing. In others, definitions of lexical items are jumbled with the intended lexical item. In light of these challenges, we initiated meticulous data-cleaning (§3.1) and expansion (§3.2) processes detailed below:

## 3.1. Dataset Cleaning

Initially, given the formulaic nature of certain puddles (e.g., the Bible), rule-based corrections enabled immediate annotation of multiple entries in some of the puddles, without subsequent cleaning. The comprehensive rules used in this phase are detailed in Appendix A.1.

For cleaning the rest, we manually corrected five-to-ten entries for each of the 131 puddles. Using ChatGPT (OpenAI, 2022), we defined a pseudo function to clean each entry in each puddle following some few-shot examples sampled from these manually corrected entries. Concretely, the pseudo function receives the number of signs (as a length indicator), language code, and existing spoken language candidates (terms) and returns a cleaned, parallel version of the terms: `clean(number of signs, language code, terms)`. An illustration would be the function call: `clean(1, "sl", ["Koreja (mednarodno)", "Korea", "S125-P1"])` returning `["Koreja", "Korea"]`. Prompts and examples are detailed in Appendix B.1.

We tested the efficacy of this cleaning method using the `gpt-3.5-turbo-0613` model on the manually cleaned test examples from each puddle. We assessed the quality via the Intersection over Union (IoU)[3] metric between the predicted terms and the annotated terms, averaged across the entire test set. We compared multiple settings, with various approaches to cleaning the data:

1. **E0:** No changes (no cleaning just copying).
2. **E1:** Rule-based cleaning (Appendix A.2).
3. **E2:** E1 + ChatGPT with four fixed, manually selected few-shot examples from various puddles.
4. **E3:** E1 + ChatGPT with up to five few-shot examples from the same puddle.
5. **E4:** E1 + ChatGPT with four fixed examples from various puddles (E2) and five examples from the same puddle (E3).
6. **E5:** E4 + using `gpt-4-0613`.

Doing nothing (*E0*) leads to a base IoU of **0.50**. The rule-based approach (*E1*), which conservatively eliminated undesired text entries, provided a slight boost, resulting in an IoU of **0.53**. Incorporating general few-shot examples into the cleaning process (*E2*) significantly increased the IoU to **0.63**. A more targeted approach using five few-shot examples from the same puddle (*E3*) further improved this to **0.71** IoU. When combining the general few-shot examples with puddle-specific examples (*E4*), we achieved an IoU of **0.74**. Our best results, however, came from GPT-4 (*E5*), which achieved an IoU of **0.80**.

---

[3] Note: The maximum IoU is not 1. We can not ignore possible human annotation errors/variations, especially when dealing with non-English data.

For cost considerations, the following pricing was assumed: $0.0015/1K$ tokens for `gpt-3.5-turbo` and $0.03/1K$ tokens for `gpt-4`, indicating a $20\times$ price disparity. Given the average of 714 tokens for *E4* and *E5* and around $200K$ annotations, the projected costs for `gpt-3.5-turbo` and `gpt-4` are approximately $200 and $4000, respectively. For financial reasons, we use `gpt-3.5-turbo`. The final cost ended up being $230.18, paid to OpenAI.

### 3.2. Dataset Expansion

Our next objective is to further enrich the dataset by introducing variations for each cleaned term. Variability in language representation can significantly benefit the robustness of machine translation models by providing multiple references for expressing the same idea. We designed another pseudo function for this, `expand(language code, terms)`, producing expanded terms and proper capitalization. As some terms were in English, outputs for both the specific language and English were generated separately. Prompt in Appendix B.2.

For an illustration, consider a term in Swedish such as 'tre' ('three' in English) passed to our function: `expand("sv", ["tre"])`, the returned output could be `{"sv": ["Tre", "3"], "en": ["Three", "3"]}`. This means that for the Swedish language ('sv'), the term 'tre' can be represented as 'Tre' or the numeral '3'. The corresponding English translation for the term would be 'Three'. Another example would be the German term 'Vater' ('father' in English). The function call `expand("de", ["Vater", "father"])` yields `{"de": ["Vater", "Vati", "Papa", "Erzeuger"], "en": ["Father", "Dad", "Daddy"]}`. Here, the term expands to multiple terms in both German and English.

This expansion approach (using `gpt-3.5-turbo` with 9 fixed few-shot examples), although seemingly straightforward with a similar cost to the cleaning process, introduces vast richness to our dataset. Each term is now associated with multiple representations, thereby enhancing the potential of our model to understand the nuances and variability of language. However, this expansion can also introduce errors, either when expanding terms that were not properly cleaned, or when the expansion itself is wrong. The expansion cost ended up being $299.72, paid to OpenAI.

Evaluating the efficacy of this expansion step is non-trivial, due to the inherent subjectivity involved in determining which expansions are valid or more useful than others. Interested readers are referred to Appendix C for more outputs.

## 4. Data Quality Experiments

To evaluate the quality of our cleaning and expansion, we test its effect on machine translation. We train machine translation models on the original data, on the cleaned data, and on the expanded data, in an imbalanced multilingual setting that contains all of the puddles on SignBank. For this comparison, we focus on the *signed-to-spoken* direction, since automatic evaluation of spoken language machine translation output is well established. For the development set, in each data scenario, we consider the first 3000 entries. For the test set, we use our manually annotated data from §3.1. In the source text, we include language tags to indicate the source and target language for the translation (Johnson et al., 2017). We use sacre-BLEU 2.3.1 (Post, 2018), to evaluate `BLEU`[4] (Papineni et al., 2002) and `chrF`[5] (Popović, 2016).

This comparison is only made to evaluate the quality of the different datasets. Thus, for every framework, we use the default training settings and avoid attempting to optimize with smaller models or different architecture. We posit that better test-set performance in a given framework indicates higher data quality. While we believe that this effect should be highly potent for the *spoken-to-signed* translation direction, it is not evaluated in this work since there are no human-validated automatic metrics to evaluate SignWriting output.

**Sockeye / Fairseq / OpenNMT** The SignWriting text is tokenized using §4.1, and the spoken language text is tokenized using BPE (Sennrich et al., 2016) with 3000 merges. For the cleaned dataset, this results in a smaller vocabulary than for the original dataset since some unigrams are filtered out. Model training is early-stopped on validation `chrF` score (Sockeye), `BLEU` (Fairseq), and accuracy (OpenNMT) with a patience of 10 epochs.

**Keras (Chollet et al., 2015)** To address the effect of clean data on pre-trained language models, we fine-tune *mT5-small* (Xue et al., 2021) using Keras and HuggingFace Transformers (Wolf et al., 2020). In this setting, both the source and target texts are tokenized using the *mT5* tokenizer. Since our source data is extremely out-of-domain to the original language model training, we do not expect to see improvements from the pre-trained language model compared to the machine translation frameworks mentioned above. The model is fine-tuned for up to 20 epochs, early stopped on validation loss.

---
[4] `BLEU` = case:mixed/eff:no/tok:13a/smooth:exp
[5] `chrF` = case:mixed/eff:yes/nc:6/nw:0/space:no

| Dataset | Training Pairs | Vocab | Sockeye | | Fairseq | | OpenNMT | | Keras (mT5) | | Vocab | Jiang et al. (2023) | |
|---|---|---|---|---|---|---|---|---|---|---|---|---|---|
| | | | BLEU | chrF | BLEU | chrF | BLEU | chrF | BLEU | chrF | | BLEU | chrF |
| Original | 521,390 | 6,016 | 0.23 | 10.01 | 0.18 | 4.74 | 0.69 | 9.21 | 0.07 | 6.39 | 3,872 | 6.44 | 22.01 |
| Cleaned | 357,574 | 5,200 | **24.33** | **27.88** | 1.1 | **7.59** | 30.6 | 22.46 | **6.02** | 12.35 | 3,752 | **24.65** | **31.22** |
| Expanded | 1,027,418 | 5,976 | 18.56 | 17.74 | **1.26** | 6.52 | 13.38 | 13.0 | 2.99 | **12.49** | 4,312 | 19.73 | 25.12 |

Table 1: Evaluation of the usability of our data for machine translation.

**Factord Machine Translation (Jiang et al., 2023)** For completeness, we additionally train the complex factored models specially designed and optimized for SignWriting translation, following experimental settings proposed by the previous work. Note that their approach generally uses a smaller vocabulary size to accommodate a low-resource machine translation scenario (Sennrich and Zhang, 2019).

### 4.1. SignWriting Tokenization

We tokenize the Formal SignWriting in ASCII (FSW) sequences, a linear format of SignWiring, into discrete tokens. For example, the American Sign Language sign for "Hello" is represented as:

`M518x529S14c20481x471S27106503x489`

This representation is a sequence of graphemes, each with a structure of a `symbol`, `modifiers`, and an `<x,y> position` (Table 2).

| | symbol | modifiers | x | y |
|---|---|---|---|---|
| 1. | M | - | 518 | 529 |
| 2. | S14c | 2 0 | 481 | 471 |
| 3. | S271 | 0 6 | 503 | 489 |

Table 2: Tokenized ASL sign "Hello".

From this structure, we treat each component as a separate token. We further remove predictable and redundant symbols, such as `M`, `x` and the size of the box, resulting in the following sequence:

`M p518 p529`
`S14c c2 r0 p481 p471`
`S271 c0 r6 p503 p489`

This tokenization process simplifies the complex FSW strings, creating a small vocabulary of 1182 tokens for our NMT framework (4 boxes, 656 symbols, 6 plane modifiers, 16 rotation modifiers, and 500 positions).

## 5. Results

Table 1 shows that despite the different frameworks, pre-trained models, unoptimized modeling, and imbalanced multilingual translation scenarios, performance on the cleaned data is consistently better compared to the original data. This establishes our cleaned data as more useful for signed-to-spoken machine translation.

Comparing the performance of naive text-to-text machine translation approaches to the factored approach proposed by Jiang et al. (2023) on the original dataset, our experimental results confirm their approach's efficiency on a raw and noisy dataset. However, the performance gap between simple and complex modeling is nearly closed when we move training from the original dataset to the cleaned dataset. Despite their optimization in modeling, our optimization in data quality makes up for simple modeling. After we expand the dataset, most models see a performance drop due to possible confusion and noise caused by multi-translation targets introduced by the expansion. In such a case, Sockeye and the factored approach remain relatively robust.

Overall, the use of our expanded data is dubious for *signed-to-spoken* machine translation. If our cleaned data is of perfectly good quality, our expansion can only add noise by introducing multiple targets for the same source. However, since we know that our cleaned data is not perfect, we hypothesize that the additional noise from the data expansion smooths out the noise in the imperfect data, by introducing more overlaps between identical translations, thus drowning the noise. This is very difficult to evaluate. As we vary the target texts in many dimensions (gender, formality, capitalization, script, and form), uncontrolled translation of the test set into the original distribution of these dimensions is improbable, even when disregarding noise coming from wrong expansions. This is reflected in the results. Using the expanded data for pre-training our Sockeye model, then fine-tuning on the cleaned data gets the model back to the target distribution, improved results of $26.07$ BLEU and $28.84$ chrF.

## 6. Conclusions

This work introduces a methodology for data cleaning and expansion for low-resource settings such as sign language translation. Its main contribution is the introduction of *SignBank+*, a cleaner and more expansive sign language translation dataset than its origin, *SignBank*. By studying the case of machine translation, we show empirically that the improvement in data quality raises equally strong performance gains as a sophisticated modeling approach does. The data and the code are publically available on https://github.com/sign-language-processing/signbank-plus.

## 7. Future Work

We encourage future work to expand on our efforts and create *SignBank++*. The *clean* and *expand* steps can be executed with more, and better language models. Quality estimation filtering methods can be created to filter out text pairs likely to not be parallel (Junczys-Dowmunt, 2018; Bane et al., 2022; Steingrímsson et al., 2023). Additionally, optimizing the input representation by encoding SignWriting as images (Dosovitskiy et al., 2021), reducing the token count, and standardizing phoneme order could all lead to improved translation performance. Finally, robust evaluation metrics for spoken-to-signed translation should be created and validated with human judgments.

## 8. Limitations

This section discusses the limitations of our study, which future research might address or consider.

**ChatGPT** The use of ChatGPT to clean and expand the SignBank dataset, while efficient, introduces certain limitations. The model's generated outputs might carry subtle biases or inaccuracies not easily discernible, potentially affecting the dataset's integrity. Our research relied on proprietary models (ChatGPT, *gpt-3.5-turbo*, and *gpt-4*) for significant processes, which might limit accessibility for some researchers due to costs or usage restrictions. This dependency also poses challenges for the exact reproducibility and scalability of our findings, since these models are no longer available.

**Evaluation** Our study lacks extensive human evaluation in assessing the cleaned and expanded dataset's quality and the translations' accuracy. Automatic metrics like BLEU and chrF, although useful, cannot fully grasp the fluency and contextual appropriateness of translations from a human perspective. The reliance on these metrics might not reflect the true quality accurately.

**Performance on Expanded Data** The dataset expansion's impact on model performance was varied, indicating that while it adds richness, it may also introduce potential noise. This underscores the challenge of balancing between enriching the dataset and maintaining its quality. Furthermore, additional research is needed to verify its effectiveness in the spoken-to-signed translation direction.

## 9. Bibliographical References

# A. Rules

## A.1. Automatic Annotation Rules

**Question Marks** It is rare, but sometimes, this movement symbol is used as a question mark, because of visual resemblance. We remove all entries that contain only a question mark (M510x517S29f0c491x484).

**Korean (puddle 78)** This large puddle ($25k$ entries) is quite standardized. Most entries include four terms, in a predictable fashion. For example, {23-6524-385267, 1,5, .} For all $22k$ entries that match this fashion, we annotate them with the second term, excluding the number that follows. For the above example, we use .

**Slovene (puddle 52)** Out of $6k$ entries, about $3k$ seem to fit a specific pattern. A single term, with possibly a single uppercase letter (variation) and the source in parenthesis. For example, {zdarma B (UPOL)} is annotated by removing the variation and source, to result in zdarma.

**The Bible (puddles 151 and 152)** These puddles include translation of the Bible into SignWriting in Signed Exact English (SEE) and not American Sign Language (ASL). Almost every entry includes a book, chapter, and verse identifier, for example 1Corinthians01v03 means The First Epistle to the Corinthians, Chapter 1, Verse 3. We only address entries that we can extract the book, chapter, and verse from, and that are of a single verse, not split apart (some entries contain parts of verses, and others contain multiple verses). Based on the match, we extract the verse from the *bible-corpus*[6], and disregard any other text in the entry. In some entries, the SignWriting starts with indicating 'Verse' and a number. We attempt to recognize when this happens based on simple string matching, and when it does, we add Verse {number}: to the beginning of the verse.

## A.2. Data Filtering Rules

- We remove all terms that include a URL in them. These usually link to an image, a video, or a source.

- For Slovene entries in puddle 52 that did not match our criteria for automatic annotation, we strip the suffix as mentioned above from all terms.

- For Swiss-French in puddle 49, we remove entries that indicate the source based on the following regex: (lexique SGBFSS|lexique SGB-FSS^|liste: |jeu SignEcriture |JEU-COULEURS |CCSS |ApéroSignes). We also remove entries that start with FMS or EMM, or include a n°

- For Honduran in puddle 16, we remove all entries that include 'SWS-TAG'.

- For German in puddle 53, we remove all entries that include vgl, KK or delegs, or that match Variante \d, Geschichte ".*?", [Ss][\d. ]* or rwth\d*.

- For American in puddle 4, we remove all entries equal to English sign.

- For Argentinian in puddle 41, we remove all entries that start with .LSC.

- For Quebecois in puddle 47, we remove all entries that start with either Liste: or Alice. We also filter out the last term if it matches a part of speech (nom, verbe, adjectif, adverbe, pronom, préposition, conjonction, interjection, déterminant, phrase, géographie).

---

[6]https://github.com/christos-c/bible-corpus

# B. Prompts

## B.1. Data Cleaning

**System**

```
You are a proficient assistant, responsible for data sanitization for a machine \
translation model. Your main task involves operating the `clean` function:

The `clean` function is specifically designed to return a list of accurate translations \
that correspond to the provided SignWriting text.
The text output must be valid spoken language text, fixed in case of errors.
This aids in the training of machine translation models.
Since SignWriting can also write fingerspelling, \
a single character can also be a valid output.

When confronted with an unfamiliar SignWriting string, \
the function employs heuristic methods, including:

1. A similarity in the number of words to the number of signs. \
For example, a string with 20 signs and 3 words is unlikely to have a parallel translation.
2. Potential multiple entries are considered when several terms in the text have the same \
meaning. For instance, "one (1)" or "one-also-1" or "one/1" will result in "one" and "1".
3. In cases of uncertainty, the function returns an empty array.

    Parameters:
        SignWriting (int): Number of signs present.
        Language (string): language code (Optional)
        texts (list): A list of raw, unfiltered candidate texts.

    Returns:
        translations (list): A list of strings that accurately match the SignWriting, \
        intended for machine translation.

In the course of the session, users will invoke the `clean` function and you will \
respond with the function's output.
```

**Manually Curated Few Shots**

| User | Assistant |
|---|---|
| `clean(1, "sl", ["Koreja (mednarodno)", "Korea"])` | `["Koreja", "Korea"]` |
| `clean(1, "sl", ["Bosna in Hercegovina 2", "Bosnia and Herzegovina"])` | `["Bosna in Hercegovina", "Bosnia and Herzegovina"]` |
| `clean(18, "en", ["Acts 04_27-31c", "James Orlow"])` | `[]` |
| `clean(8, "es", ["Juan el Bautista predica", "1:1 El principio de la buena noticia de Jesucristo, el Hijo de Dios."])` | `["El principio de la buena noticia de Jesucristo, el Hijo de Dios."]` |

Table 3: Manually curated few-shot `user` and `assistant` interactions.

## B.2. Data Expansion

**System**

```
You are a proficient assistant, responsible for data augmentation for a \
machine translation model. Your main task involves executing the `expand` function:

The expand function is specifically crafted to return a list of equivalent \
expressions that match the provided text in a certain language. \
The text output must be accurately spelled and grammatically correct spoken \
language text. This is beneficial for improving the robustness of \
machine translation models.

The function uses a series of methods to ensure a variety of equivalent terms, \
including:

1. Paraphrases in the same language. For example, the text "hello" can be \
paraphrased as "Hi" or "Hey"
2. Capitalization corrections: For instance, the text "one" may result in ["One"]. \
"donald duck" may result in ["Donald Duck"]
3. Numerical translation: When the text represents a number, its numeric \
equivalent is added. For example, "one" would return ["One", "1"].
4. Language translation: If the text is in a non-English language, and it's feasible \
to translate it to English, the English translation is included. For instance, \
"domingo" would return ["Domingo", "Sunday"].
5. If the text is not spoken language text, but instead a random identifier \
like "rom-ale-10-44r", ignore it completely.

  Parameters:
    language (string): language code for the terms in the list.
    texts (list): A list of terms requiring expansion.

  Returns:
    expansions (obj):
      language (list): A list of unique strings that are equivalent to the \
      provided terms, intended for machine translation training.
      en (list): A list of unique translations to English if the language is \
      not english and translation is feasible.

During the session, users will call the `expand` function and you \
will respond with the function's output.
```

### Manually Curated Few Shots

| User | Assistant |
| --- | --- |
| `expand("sl", ["2"])` | `{"sl": ["2", "Dva"], "en": ["2", "Two"]}` |
| `expand("fr", ["Lac", "LEMAN"])` | `{"fr": ["Lac Leman"], "en": ["Lake Geneva"]}` |
| `expand("no", ["fire", "tall", "4"])` | `{"no": ["fire", "4"], "en": ["four", "4"]}` |
| `expand("he", ([" "]` | `{"he": ," "] ,[" " "en": ["Aleph", "Alef"]}` |
| `expand("pt", ["MAGIA", "MAGICO"])` | `{"pt": ["MAGIA", "MÁGICO"], "en": ["Magic", "Magical", "Magician"]}` |
| `expand("de", ["Vater", "father"])` | `{"de": ["Vater", "Papa", "Papi", "Vati", "Erzeuger"], "en": ["Father", "Dad", "Daddy", "Papa"]}` |
| `expand("en", ["Hello", "hi"])` | `{"en": ["Hello", "Hi", "Hey", "Greetings", "Howdy", "Hiya", "Aloha", "Bonjour", "Hola", "Salutations", "Hello there", "Hi there"]}` |
| `expand("pt", ["Eu-tentar"])` | `{"pt": ["Eu tentar"], "en": ["I try"]}` |
| `expand("de", ["zwaiund zwanzig", "S3-07163-V"])` | `{"de": ["Zwei und Zwanzig", "22", "Zweiundzwanzig"], "en": ["Twenty-two", "22"]}` |

Table 4: Manually curated few-shot `user` and `assistant` interactions.

## C. Dataset Preview

Here we include a preview of the dataset, including the original spoken language terms, the automatically cleaned terms based on the original texts, the human annotation based on the original text, and the expanded texts based on the human annotation. We include one sample per puddle (as long as the puddle includes a sample that is less than 200 characters).

| pid | eid | Lang | Original | Cleaned (from Original) | Annotation | Expanded (from Annotation) |
|---|---|---|---|---|---|---|
| 4 | 101 | en | [glasses] | [glasses] | [Glasses] | [Glasses, Eyeglasses, Spectacles, Optical frames] |
| 5 | 296 | en | [gesture 1, heck with it] | [heck with it, gesture 1] | [heck with it] | [Heck with it, Forget it, Never mind, Screw it, Drop it, Let it go] |
| 11 | 92 | en | [cookie, biscuit, https://www.youtube.com/watch?v=jyOh9Ss7Dzs] | [cookie, biscuit] | [cookie, biscuit] | [Cookie, Biscuit, Cracker, Snack, Treat, Sweet, Baked good] |
| 13 | 2 | zh-CN | [epigrafe] | [] | [] | [] |
| 16 | 2829 | es | [soñar, SWS-TAG[SignLists:Dict Sept15 Honduras,SignLists:New Dict,Dibujos:Dibujos Grupo1]] | [soñar] | [soñar] | [soñar, soñando] |
| 17 | 2 | en | [CBS 1, Deaf Hands Story CBS 1 (as retold by Stuart)] | [CBS 1, Deaf Hands Story CBS 1] | [] | [] |
| 18 | 62 | am | [Birr, Ethiopian Currency] | [Birr] | [Birr, ] | [Birr, ] |
| 19 | 1947 | pl | [sobie, siebie, to oneself; oneself] | [to oneself, oneself, siebie, sobie] | [sobie, siebie, to oneself, oneself] | [sobie, siebie] |
| 20 | 49 | fr | [amygdalectomie] | [amygdalectomie] | [amygdalectomie] | [Amygdalectomie] |
| 21 | 624 | en | [VetDr04] | [] | [] | [] |
| 22 | 8 | fr | [ADN, ADN-définition] | [ADN] | [] | [] |
| 23 | 17 | no | [Gullhår 12, Vi tre går en tur i skogen.] | [Vi tre går en tur i skogen.] | [Vi tre går en tur i skogen.] | [Vi tre går en tur i skogen.] |
| 24 | 4 | no | [få (ikke mange)] | [få] | [få, ikke mange] | [få, ikke mange] |
| 25 | 1268 | en | [wrong, accidental, by mistake] | [by mistake, accidental, wrong] | [wrong, accidental, by mistake] | [Wrong, Incorrect, Inaccurate, Mistaken, Accidental, Unintentional, By mistake, By accident, In error] |
| 26 | 261 | de | [Städte Teil 2, Spiel: "Stadt, Land, Fluss"] | [Spiel: Stadt, Land, Fluss, Städte Teil 2] | [] | [] |
| 27 | 14 | de | [Beispielsatz index, Thomas kauft ein Auto. Es ist billig.] | [Thomas kauft ein Auto. Es ist billig.] | [Thomas kauft ein Auto. Es ist billig.] | [Thomas kauft ein Auto. Es ist billig.] |
| 28 | 11584 | en | [455] | [455] | [455] | [455, Four hundred fifty-five] |
| 29 | 4 | de | [Vater, father] | [Vater, father] | [Vater, father] | [Vater, Papa, Papi, Erzeuger] |
| 30 | 30 | da | [R, fingerspelling] | [R] | [R] | [R] |
| 31 | 173 | mt | [Kugin, Cousin, Dizzjunarju ta' Affarijiet ta' Kuljum\n\nVolum: FAMILJA] | [Cousin, Kugin] | [Kugin, Cousin] | [Kugin, Kuzin] |
| 32 | 9 | en | [God is with us, The biblical meaning of Emmanuel] | [God is with us, Emmanuel] | [God is with us] | [God is with us] |
| 33 | 409 | pt | [expressão-facial, expressão-facial] | [expressão-facial] | [expressão-facial, expressão-facial] | [Expressão facial, Rosto, Expressão no rosto] |
| 34 | 372 | th | [ , association, noun, https://www.th-sl.com/word/10-039/] | [ , association] | [ , association] | [ , ] |
| 35 | 6 | en | [Arkansas, US State, (n) a state in the United States.] | [Arkansas] | [Arkansas] | [Arkansas] |
| 36 | 156 | cs | [O perníkové chaloupce 2] | [O perníkové chaloupce 2] | [] | [] |
| 37 | 9 | cs | [podtřída] | [] | [] | [] |
| 38 | 4 | pl | [Miraks i Mariks 1, Ja muszę mówić cicho, ponieważ moi ludzie nie mogą mnie słyszeć.] | [Ja muszę mówić cicho, ponieważ moi ludzie nie mogą mnie słyszeć.] | [Ja muszę mówić cicho, ponieważ moi ludzie nie mogą mnie słyszeć.] | [Ja muszę mówić cicho, ponieważ moi ludzie nie mogą mnie słyszeć.] |
| 40 | 277 | ar | ] [ | , ] , [ | [] | [] |
| 41 | 1765 | es | [Grace, .LSC vocab Personas] | [Grace] | [Grace] | [Grace] |
| 42 | 23 | en | [brother] | [brother] | [brother] | [Brother, Sibling, Bro, Buddy, Mate, Pal, Comrade, Fellow] |
| 43 | 1396 | fr | [cinéma] | [cinéma] | [cinéma] | [cinéma] |
| 44 | 5617 | nl | [Jelle, jelle] | [jelle, Jelle] | [Jelle, jelle] | [Jelle] |
| 45 | 3 | es | [walk] | [caminar] | [walk] | [caminar, andar, pasear] |
| 46 | 11173 | pt | [alfabeto] | [alfabeto] | [] | [] |
| 47 | 10094 | fr | [trésorier, trésorière, nom] | [trésorier, trésorière] | [trésorier, trésorière] | [trésorier, trésorière] |
| 48 | 3891 | de | [glcklich-2] | [glcklich] | [Glücklich, Happy] | [Glücklich, Froh, Fröhlich, Zufrieden] |
| 49 | 1267 | fr | [jeu SignEcriture, 3-11-4] | [] | [] | [] |
| 50 | 2 | it | [VAUD, canton Suisse] | [VAUD] | [VAUD] | [VAUD] |
| 51 | 801 | es | [pensamiento] | [pensamiento] | [pensamiento] | [pensamiento, reflexión, idea, concepción, cogitación] |
| 52 | 1007 | sk | [displej (IMoTeSP)] | [displej] | [displej] | [displej, obrazovka] |
| 53 | 12013 | de | [für, hier: für 2010 (obewohl kein Bonativ?)] | [für, hier: für 2010] | [für] | [für] |
| 54 | 1222 | eo | [ist, G@17] | [ist] | [ist] | [estas, estis, estos] |
| 55 | 34 | es | [tocar] | [tocar] | [tocar] | [tocar, reproducir, interpretar, ejecutar] |
| 56 | 1605 | ca | [geografía] | [geografía, geografia, geography] | [geografía] | [geografia, geografies] |
| 57 | 173 | fi | [TAVATA] | [TAVATA] | [TAVATA] | [TAVATA, Tavata] |
| 58 | 1375 | fr | [surnom] | [surnom] | [surnom] | [surnom, sobriquet, pseudo, surnommer] |
| 59 | 274 | en | [boy, Theme: family details, son, theme: cards] | [boy, son] | [boy, son] | [Boy, Son, Child, Youngster, Lad, Kid, Offspring, Male child] |
| 60 | 12 | en | [seven] | [seven] | [seven] | [Seven, 7] |
| 61 | 20 | gr | [sigma, Eighteenth Letter in the Greek Manual Alphabet] | [sigma, Σ, σ] | [sigma, Σ] | [sigma, Σ] |
| 62 | 35 | en | [A, fingerspelling] | [A] | [A] | [A, One] |
| 63 | 677 | it | [(passato) incontrare] | [incontrare] | [incontrare] | [incontrare, incontrarsi] |
| 64 | 227 | ja | [ , , ki, JSL Fingerspelling] | [ , , ki] | [ , , ki] | [ , , , ] |
| 65 | 65 | es | [Uno, Uno (one)] | [Uno, one] | [Uno, one] | [Uno, 1] |
| 66 | 3 | ms | [Malaysia] | [Malaysia] | [Malaysia] | [Malaysia] |
| 67 | 1598 | es | [banarse, bathe] | [bañarse] | [banarse, bathe] | [bañarse, ducharse] |
| 68 | 68 | nl | [AANGIFTE / AANGEVEN ()] | [AANGIFTE, AANGEVEN] | [AANGIFTE, AANGEVEN] | [AANGIFTE, AANGEVEN, Melding, Melden, Aangifte doen] |
| 69 | 679 | no | [4, fire, tall] | [fire, tall, 4] | [4] | [4, Fire] |
| 70 | 33 | en | [I] | [I] | [I] | [I, Me, Myself] |
| 71 | 67 | es | [flor, flower] | [flor, flower] | [flor, flower] | [flor, flora] |
| 72 | 325 | fil | [mountain] | [mountain] | [mountain] | [bundok] |
| 73 | 72 | sv | [Ä, fingerspelling] | [Ä, ä] | [Ä] | [Ä, A med ring över] |
| 74 | 2477 | sl | [pojutrišnjem, day after tomorrow] | [pojutrišnjem, day after tomorrow] | [pojutrišnjem, the day after tomorrow] | [pojutrišnjem, pojutrišnjem dnevu] |
| 75 | 6 | zh-tw | [girl] | [girl, ] | [] | [] |
| 76 | 39 | es | [y] | [y] | [y] | [y, e] |
| 77 | 45 | en | [Science] | [Science] | [Science] | [Science, Scientific, Sciences] |
| 78 | 10000 | ko | [ _0216, 1, 6, .] | [ ] | [ ] | [ , ] |
| 79 | 10 | sw | [see] | [see] | [see] | [see] |
| 80 | 883 | pt | [barriga] | [barriga] | [Barriga, Stomach] | [Barriga, Estômago] |
| 81 | 36 | fr | [Les cinq frères chinois 02] | [] | [] | [] |
| 82 | 115 | sq | [Itali, italy] | [Itali, Italy] | [Itali, italy] | [Itali, Italia] |
| 83 | 98 | zh-cn | [7, , number] | [7, number, ] | [7, ] | [7, , ] |
| 84 | 54 | ar | , ] , , , water, [ | , ] , , , water] | , ] , , , water, [ | , ] , , , [ |
| 85 | 2 | hi | [Indian (from India)] | [Indian] | [Indian] | [ , ] |
| 86 | 17 | ar | , ] [ | , ] [ | [ ] | , ] [ |
| 87 | 3 | ur | [school] | [school] | [school] | [ ] |
| 88 | 42 | ru | [ , 7] | [ , 7] | [ , 7] | [ , 7] |
| 89 | 260 | sk | [chlapec] | [chlapec] | [chlapec] | [chlapec, chalan, kluk, chlapčenský] |
| 90 | 34 | tr | [o, he/she/it; er/sie/es] | [o, he/she/it, er/sie/es] | [o, he, she, it, er, sie, es] | [o, he, she, it, er, sie, es] |
| 91 | 28 | ar | [Goldilocks Page 1] | [] | [] | [] |
| 92 | 1 | ar | ] [ | ] [ | ] [ | ] [ |
| 93 | 527 | es | [SignoEscritura Reading Lessons pg. 20] | [] | [] | [] |
| 94 | 25 | ca | [patinatge artístic] | [patinatge artístic] | [patinatge artístic] | [Patinatge artístic, Patinatge artístic sobre gel] |
| 96 | 13 | de | [Noah 15] | [] | [] | [] |
| 98 | 35 | nl | [android] | [android] | [android] | [Android] |
| 99 | 16 | ja | [texto11] | [] | [] | [] |
| 100 | 1 | am | [texto11] | [texto11] | [] | [] |
| 103 | 51 | mt | [Luqa1v10, Waqt ir-rit ta' l-incens, il-kotra kollha tal-poplu kienet qieghda titlob barra.] | [Waqt ir-rit ta' l-incens, il-kotra kollha tal-poplu kienet qieghda titlob barra.] | [Waqt ir-rit ta' l-incens, il-kotra kollha tal-poplu kienet qieghda titlob barra.] | [Waqt ir-rit ta' l-incens, il-kotra kollha tal-poplu kienet qieġħda titlob barra.] |
| 104 | 1181 | ar | [Bonjour] | [Bonjour] | [] | [] |
| 105 | 338 | en | [exit, leave, out] | [exit, out, leave] | [exit, leave, out] | [Exit, Leave, Out, Quit, Depart, Go away, Withdraw, Egress, Vacate] |
| 106 | 2 | ps | [mother, mAdar, [ | [mother, ,    mAdar] | [mother, mAdar, [ | [mother, [ |
| 110 | 15 | he | , ] fingerspelling] | [ ] | [ ] | , ] [ |
| 111 | 192 | en | [why] | [why] | [why] | [Why, For what reason, What is the reason, What is the purpose, What is the cause] |
| 113 | 1 | ht | [Zebra] | [Zebra] | [Zebra] | [Zebra] |
| 114 | 2190 | pt | [prova] | [prova] | [prova] | [prova, teste, exame] |
| 115 | 24 | pt | [bebe12] | [] | [] | [] |
| 116 | 1348 | pt | [Dentes superiores tocando a língua] | [Dentes superiores tocando a língua] | [] | [] |
| 117 | 8 | pt | [Isabel Morais, Nome Gesto] | [] | [] | [] |
| 119 | 404 | es | [World Explorers Part 1 pg. 06] | [] | [] | [] |
| 120 | 17 | es | [Tapa abriendo] | [Tapa abriendo] | [Tapa abriendo] | [Tapa abriendo] |
| 122 | 3 | hu | [Ország, Country, Land] | [Ország, Country, Land] | [Ország, Country, Land] | [Ország, Országok] |
| 123 | 1 | hu | [raus, im ärgerlichen Befehlston] | [raus] | [raus] | [raus] |
| 124 | 15 | fr | [corbeau] | [corbeau] | [corbeau] | [corbeau, corneille, corbin] |
| 125 | 16 | en | [The farmer is in his den, part 6, Theme: practice makes perfect DVD, The dog wants a bone, the dog wants a bone, E I A D O, the dog wants a bone.] | [The farmer is in his den, part 6, The dog wants a bone, the dog wants a bone, E I A D O, the dog wants a bone.] | [The dog wants a bone, the dog wants a bone, E I A D O, the dog wants a bone.] | [The dog wants a bone, the dog wants a bone, E I A D O, the dog wants a bone.] |
| 126 | 319 | ar | [Luc4:23 (LSF)] | [] | [] | [] |
| 127 | 1171 | mt | [Heading 10, Il-ħarba lejn l-Eġittu] | [Il-ħarba lejn l-Eġittu] | [Il-ħarba lejn l-Eġittu, The flight to Egypt] | [Il-ħarba lejn l-Eġittu] |
| 128 | 41 | mw | [teacher] | [teacher] | [teacher] | [teacher] |
| 129 | 5284 | gn | [Mejorar 2] | [Mejorar, 2] | [Mejorar] | [Mejorar] |
| 130 | 27 | uk | [ , ] | [ , ] | [ , ] | [ , ] |
| 131 | 36 | is | [stór (pf.1)] | [stór] | [stór] | [stór, stórt] |
| 132 | 12 | ro | [luni, Transilvania Semne, Monday, Montag] | [Monday, Montag, luni] | [luni, Monday, Montag] | [luni] |
| 133 | 3 | ne | [is, chha, AAMA CHHA? (do you have a mother?)\nthe copula used in a more restricted way in NSL than in spoken Nepali] | [is, chha] | [chha] | [ , ] |
| 134 | 1 | bg | [    , hello] | [    , hello] | [    , hello] | [    ,     ,     ,     ] |
| 135 | 89 | es | [h, Letra (consonante)] | [h] | [h] | [h] |
| 137 | 14 | es | [San Salvador, san salvador, Capital (El Salvador), Geografia.] | [San Salvador] | [San Salvador, san salvador] | [San Salvador] |
| 143 | 84 | es | [abierto] | [abierto] | [abierto] | [abierto, desbloqueado, libre, sin restricciones] |
| 147 | 1154 | mt | [Taken from...Ghaqda, Mehud mit-test tal-Ghaqda Bibblika ..etc] | [Mehud mit-test tal-Ghaqda Bibblika ..etc] | [] | [] |
| 148 | 17 | sl | [Sveti Filip, prosi za nas!] | [Sveti Filip, prosi za nas!] | [Sveti Filip, prosi za nas!] | [Sveti Filip, prosi za nas!] |
| 151 | 10009 | en | [Matthew15v07 NLT, You hypocrites! Isaiah was right when he prophesied about you, for he wrote,\n\nMatthew15v7 NLT] | [Verse 7: You hypocrites! Well did Isaiah prophesy of you, saying:, Verse 7: Ye hypocrites, well did Esaias prophesy of you, saying,] | [Verse 7: You hypocrites! Well did Isaiah prophesy of you, saying,, Verse 7: Ye hypocrites, well did Esaias prophesy of you, saying,] | [] |
| 152 | 10173 | en | [Proverbs 24:28, Pr 24:28 ¶ Be not a witness against thy neighbour without cause; and deceive [not] with thy lips.\n\nDo not witness against neighbor for no reason and do not deceive people.] | [Be not a witness against your neighbor without cause; and deceive not with your lips., Do not witness against neighbor without cause; and deceive not with your lips.] | [Don't be a witness against your neighbor without cause. Don't deceive with your lips., Be not a witness against thy neighbour without cause; and deceive not with thy lips.] | [] |
| 153 | 448 | vn | [A cho B, A gives to B something, https://www.youtube.com/watch?v=9trjbdhheIo] | [A cho B, A gives to B something] | [cho, gives] | [cho, đưa] |